\documentclass{ecai} 

\usepackage{latexsym}
\usepackage{amssymb}
\usepackage{amsmath}
\usepackage{amsthm}
\usepackage{booktabs}
\usepackage{enumitem}
\usepackage{graphicx}
\usepackage{color}

\newtheorem{theorem}{Theorem}

\newtheorem{definition}{Definition}

\usepackage{mathtools} 
\newtheorem{example}{Example}

\DeclareMathOperator*{\argmax}{\arg\!\max} 
\newcommand{\norm}[1]{\left\lVert #1 \right\rVert} 

\DeclarePairedDelimiterX{\infdivx}[2]{(}{)}{%
  #1\;\delimsize\|\;#2%
}

\newcommand{\BibTeX}{B\kern-.05em{\sc i\kern-.025em b}\kern-.08em\TeX}

\begin{document}


\begin{frontmatter}


\paperid{123} 


\title{(Un)certainty of (Un)fairness: Preference-Based Selection of Certainly Fair Decision-Makers}


\author[A]{\fnms{Manh Khoi}~\snm{Duong}\orcid{0000-0002-4653-7685}\thanks{Corresponding Author. Email: duong@hhu.de.}}
\author[A]{\fnms{Stefan}~\snm{Conrad}\orcid{0000-0003-2788-3854}}

\address[A]{Heinrich Heine University, Universit\"atsstra\ss{}e 1, 40225 D\"usseldorf, Germany}


\begin{abstract}
    Fairness metrics are used to assess discrimination and bias
    in decision-making processes across various domains,
    including machine learning models and human decision-makers
    in real-world applications.
    This involves calculating the disparities between probabilistic outcomes
    among social groups, such as acceptance rates between male and
    female applicants. 
    However, traditional fairness metrics do not account for the
    uncertainty in these processes and lack of comparability when
    two decision-makers exhibit the same disparity.
    Using Bayesian statistics, we quantify the uncertainty of the disparity
    to enhance discrimination assessments.
    We represent each decision-maker, whether a machine learning model
    or a human, by its disparity and the corresponding uncertainty
    in that disparity.
    We define preferences over decision-makers and utilize brute-force to
    choose the optimal decision-maker according to a utility function
    that ranks decision-makers based on these preferences.
    The decision-maker with the highest utility score can be
    interpreted as the one for whom we are most certain
    that it is fair.
\end{abstract}

\end{frontmatter}


\section{Introduction}\label{section:introduction}
Traditional fairness metrics have played an important role in quantifying
disparities between different social groups in data, machine learning
predictions, and decision-making systems~\citep{liobait2017MeasuringDI, zafar2017-disparate, corbett2017conditionalstat, barocas-hardt-narayanan2019book}.
However, they fail to address the inherent uncertainty present in
real-world data, i.e., \emph{aleatoric uncertainty},
particularly when minorities or generally data samples
are underrepresented.
Our work is motivated by comparing machine learning models
regarding their fairness in any socially responsible application.
We use the umbrella term \emph{decision-maker}
which can refer to any system or human that makes decisions based on data.
Therefore, our work deals with both human and algorithmic decision-makers
and is not limited to either of them.
Still, for simplicity, our examples only involve humans.

\begin{figure}
  \centering
  \includegraphics[width=0.48\textwidth]{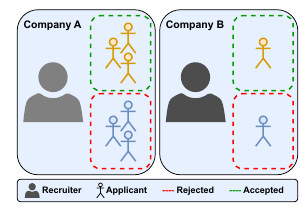}
  \caption{Both companies, $A$ and $B$,
  discriminate against applicants by only accepting yellow candidates
  and rejecting all blue candidates.
  The statistical disparity score for both companies is 100\%.
  Nevertheless, company $B$ received fewer applicants,
  making its case more uncertain.
  Which decision-maker do we favor regarding
  fairness in such a case?
  What if both companies accepted all applicants?
  Who do we favor then?}
  \label{fig:motivation}
\end{figure}

We consider an illustrative scenario (see Figure~\ref{fig:motivation})
in a hiring setting
in which two different companies, labeled $A$ and $B$, sought to hire
applicants. We also assume that all applicants in this scenario
have equal qualifications and do not differ in any way except for the
social group they belong to. Company $A$ notably only accepted
yellow candidates and rejected all blue candidates.
Company $B$ acted in the same way but received
significantly fewer applications.
When using
\emph{statistical disparity}~\citep{duan2008disparities, calders2009}
to assess discrimination from both companies, we obtain
the same score, which is 100\%, signifying the disparity
of the chances between yellow
and blue candidates of getting accepted.
Intuitively, we are more certain about the decisions being made by
company $A$ than company $B$.
In the case of company $B$, the rejection of blue candidates
can be attributed to random circumstances.
In this case, we would judge company $A$ as more
discriminatory than company $B$
because we are more certain that $A$ is unfair and very
uncertain about the unfairness of $B$.
But if both companies accepted all applicants, the disparity would be 0\%,
and we would conversely judge $B$ as more discriminatory than $A$.
This is because we are certain that $A$ is fair, while we are
uncertain about the fairness of $B$.
Lastly, when comparing between uncertain fair and uncertain
unfair decision-makers, we would prefer the former over the latter.
These examples underscore the importance of quantifying and
assessing uncertainty in discrimination evaluations.

In the context of this example, we use the notation $A \succ B$ to
signify a preference relation, indicating that company $A$ is
preferred over company $B$. The preferences we obtain are as follows:
\begin{align}
	\text{fair certain} &\succ \text{fair uncertain}, \label{eq:preferences1} \\
	\text{fair uncertain} &\succ \text{unfair uncertain}, \\
	\text{unfair uncertain} &\succ \text{unfair certain}, \label{eq:preferences3}
\end{align}
where unfair and fair refer to a disparity of 100\% and 0\%, respectively.
With these \emph{trivial preferences}, following \emph{research questions} arise:
\begin{itemize}
  \item \textbf{RQ1:} How do we quantify the uncertainty of a decision-maker's (un)fairness?
  \item \textbf{RQ2:} How do we compare decision-makers that exhibit different levels of disparity and uncertainty on a continuous scale? How do we express preferences over them and rank them accordingly?
  \item \textbf{RQ3:} How do we select the optimal decision-maker according to our preferences?
\end{itemize}
We note that the task of selecting the most preferable
decision-maker cannot be done by determining the Pareto front
because uncertain cases can seem more or less fair than certain cases
depending on the circumstances.
This can be observed in the preferences (2) and (3).
Furthermore, disparity and uncertainty are not necessarily
discrete values, making it \emph{non-trivial} to compare
between decision-makers that are represented by them.
To answer the research questions, our paper's structure and contributions are as follows:
\begin{itemize}
  \item We first introduce a notation generalizing various group
  fairness criteria, eliminating the limitation to a particular group fairness criterion in our work.
  \item Using our notation, we demonstrate how to quantify uncertainties
  of group disparities exhibited by decision-makers using
  \emph{Bayesian statistics}~\citep{gelman1995bayesian} (\textbf{RQ1}).
  \item Representing decision-makers by their disparate treatments of
  groups and the uncertainty of it,
  we formally define preferences over decision-makers (\textbf{RQ2}).
  By introducing a utility function, that assigns a value to each decision-maker, we are able to
  select the optimal decision-maker from a set of decision-makers (\textbf{RQ3}). The utility values allow ranking decision-makers according
  to our preferences.
  \item We evaluate our methodology on synthetic and real-world datasets
  to demonstrate its practical usability and necessity.
  \item We draw ethical conclusions
	by discussing the implications of our work and the importance of
	incorporating uncertainty in discrimination assessments.
  \item The implementation of the proposed scores and experiments can be found at
  \url{https://github.com/mkduong-ai/fairness-uncertainty-score}.
\end{itemize}

\section{Related Work}\label{section:relatedwork}
Fairness metrics have been widely studied in the
literature~\cite{liobait2017MeasuringDI,zafar2017-disparate, corbett2017conditionalstat,barocas-hardt-narayanan2019book}
and have been used to assess discrimination in various domains.
Common are group fairness metrics that
report \emph{disparities} between social groups.
In its simplest form, the disparity is calculated as the
(absolute) difference
between the outcomes of two groups, e.g., the acceptance
rates between males and females in a job application process~\cite{calders2009}.
Any other probabilistic outcome can be used as well~\cite{duan2008disparities}.
It gets more complex when more than two groups are involved.
In this case, aggregating pairwise differences~\cite{liobait2017MeasuringDI,duong2024framework} or similarly
using \emph{meta-metrics}~\cite{lum2022debiasing} are common approaches.
For example, the maximum disparity possible can be reported
in such a case~\cite{liobait2017MeasuringDI}.
Depending on the aggregation method, the intended \emph{social welfare}
is different~\cite{duong2024framework}.
When dealing with multiple protected attributes,
\emph{subgroups} (white male, black woman, etc.) can be formed by the cross-product of the protected attributes.
However, exponentially many subgroups can be formed in this way,
and any classifier can be accused of discriminating against some subgroup.
To prevent this, \citet{pmlr-v80-kearns18a} proposed to ignore subgroups that represent a small fraction of the population.
\citet{fouldsbayesian2020} generally criticized ignoring small
subgroups, as minorities are often vulnerable to discrimination.
Still, both works~\cite{pmlr-v80-kearns18a, fouldsbayesian2020}
employ a disparity calculation to measure fairness. 
Hence, the disparity serves as a base for discrimination assessment
in several problem settings.

However, relying solely on the disparity to assess discrimination can be
problematic.
Such a measurement can be uncertain, for instance, when samples
underrepresent a population due to data sparsity~\cite{ji2020uncertainty,rastogi2024uncertainty,ganesh2023impact}.
\citet{lum2022debiasing} showed that meta-metrics are statistically biased
upwards when more groups are involved.
This effect is attributed to the increased number of comparisons between groups, which raises the likelihood of observing greater disparities.
The authors combated this by deriving a correction term to \emph{de-bias} the disparity.
\citet{fouldsbayesian2020} addressed a similar problem where they
used a Bayesian approach to estimate the fairness of
underrepresented, small subgroups.
They did it as follows:
Subgroups are essentially intersections of protected attributes,
which can be represented by joint and marginal distributions.
Under the frequentist perspective, empirical counts
can be used to estimate such probabilities.
This comes with disadvantages when the counts are small
or when the subgroups are not present in the data.
In such cases, the estimates are uncertain or undefined
due to division by zero.
\citet{fouldsbayesian2020} proposed learning the marginal distribution
with probabilistic models, allowing for uncertainty quantification.
\citet{singh2021uncertainty} and~\citet{tahir2023aleatoric}
shared a similar concern about uncertainty in fairness assessments
and argued that uncertainty can lead to unfairness.

We follow a similar strategy to~\citet{fouldsbayesian2020} in our work.
We differ by allowing for a more general uncertainty quantification
of fairness that is not limited to subgroups. Additionally,
our uncertainty measure is normalized to ensure comparability.
Upon quantifying the uncertainty, we express preferences over
pairs of disparities and uncertainties, which is not done in the work by~\citet{fouldsbayesian2020}.
Further, we combine the disparity and uncertainty into a single
utility score, allowing for a straightforward comparison
and ranking of decision-makers. The ranking
reflects preferences over decision-makers, which we introduce in this work.


\section{Preliminaries}\label{section:preliminaries}
Protected attributes such as ethnicity, nationality, and gender make individuals
vulnerable to discrimination.
We define $Z$, which represents a protected attribute,
as a discrete random variable that can take on
values from the set $g$. We refer to $g$ as groups that are distinct categories an
individual can belong to.
For example, let $Z$ represent the gender, then $g$ is a set containing the genders male, female, and non-binary.
Further, $Y$ denotes the outcome of an individual, which is a binary random variable and $\hat{Y}$ is the predicted outcome.
For both, $Y$ and $\hat{Y}$, we use the values 1 and 0 to indicate positive and negative outcomes, respectively.
We define $E_1$ and $E_2$ as events, which are subsets of the sample space $\Omega$.
The sample space $\Omega$ is the set of all possible outcomes of an experiment.

With the intention of avoiding limitations on a particular
group fairness criterion, we introduce a generalized framework
through the following definition:

\begin{definition}[Treatment]\label{def:treatment}
    We refer to the conditional probability of $E_1$ given that $E_2$
    occurs and $Z$ takes on the value $i \in g$, i.e.,
    \begin{equation}
        P(E_1 \mid E_2, Z=i), \nonumber
    \end{equation}
    as the \emph{treatment} of group $i$.
\end{definition}

Using this notation, we can generalize group fairness notions
that are based on conditional probabilities,
including statistical parity, equality of opportunity,
predictive parity etc.~\citep{calders2009, hardt2016equality, zafar2017-disparate}.
These criteria typically demand equal outcomes $E_1$
for different groups $i, j \in g$ given the same events $E_2$.
Expressed with our notation, we yield:
\begin{equation}\label{eq:fairTreatment}
	P(E_1 \mid E_2, Z=i) = P(E_1 \mid E_2, Z=j).
\end{equation}
In our case, $E_1$ often represents a dichotomous outcome,
such as $Y=1$ or $Y=0$.
Therefore, $P(E_1 \mid E_2, Z=i)$ can be interpreted
as the success probability of a Binomial distribution.
In the following, we demonstrate examples of common group fairness
criteria expressed with our notation by only
specifying $E_1$ and $E_2$.

\begin{example}[Statistical Parity~\citep{calders2009}]
    Statistical parity
    requires equal positive outcomes between groups:
    \begin{equation}
        P(Y=1 \mid Z=i) = P(Y=1 \mid Z=j), \nonumber
    \end{equation}
    where $i, j \in g$ represent different groups.
    To equivalently express it with our notation, we set
    $E_1 \coloneqq (Y=1)$ and $E_2 \coloneqq \Omega$.
    By setting $E_2$ equal to the sample space,
    we compare the probabilities of the event $Y=1$ across
    different groups without conditioning on any additional event.
\end{example}

\begin{example}[Equality of Opportunity~\citep{hardt2016equality}]
    To achieve equality of opportunity, we have to set $E_1 \coloneqq (\hat{Y}=1)$ and $E_2 \coloneqq (Y=1)$, which results in:
    \begin{equation}
        P(\hat{Y}=1 \mid Y=1, Z=i) = P(\hat{Y}=1 \mid Y=1, Z=j). \nonumber
    \end{equation}
    This is equivalent to equal true positive rates across groups.
\end{example}

\begin{example}[Predictive Parity~\citep{zafar2017-disparate}]
    Predictive parity aims for equal predictive accuracy across different groups. To achieve this with our notation, we set $E_1 \coloneqq (Y=1)$, $E_2 \coloneqq (\hat{Y}=1)$ and yield:
    \begin{equation}
        P(Y=1 \mid \hat{Y}=1,  Z=i) = P(Y=1 \mid \hat{Y}=1, Z=j). \nonumber
    \end{equation}
    This is equivalent to equal positive predictive values across groups.
\end{example}

Because achieving equal probabilities for
certain outcomes is not always possible due to variations in sample sizes
in the groups, it is common to yield unequal probabilistic outcomes,
even when the outcomes are similar.
Hence, existing literature~\citep{liobait2017MeasuringDI}
use the absolute difference between the probabilities to quantify
the strength of discrimination.

\begin{definition}[Disparity]\label{eq:diff}
    We define the difference between the treatments of the groups $i, j \in g$ in the following:
    \begin{equation}
        \delta_{Z}(i, j, E_1, E_2) = |P(E_1 \mid E_2, Z=i) - P(E_1 \mid E_2, Z=j)|, \nonumber
    \end{equation}
    and refer to it as \emph{disparity}.
    The \emph{disparity} satisfies all properties of a mathematical metric regarding $i, j$ and is also referred to as \emph{fairness metric}.
\end{definition}
Higher differences indicate increased discrimination.
Trivially, $\delta_{Z}$ is commutative regarding $i, j$.
Establishing $\delta_{Z}$ provides a fundamental foundation for various scenarios.
For instance, it allows us to aggregate pairwise differences between groups,
particularly when dealing with attributes that are non-binary~\citep{liobait2017MeasuringDI,duong2023rapp,duong2023framework,duong2024measuring}.
\section{Quantifying Uncertainty}\label{section:uncertainty}
As shown in Equation~\eqref{eq:fairTreatment},
we can describe fairness criteria by demanding equal treatments.
However, the treatment of a group $i \in g$
can often exhibit uncertainty due to the limited number of samples.
In this section, we contrast frequentist and Bayesian approaches
to estimate the treatment probabilities $P(E_1 \mid E_2, Z=i)$.
We then model the uncertainty of the disparity $\delta_{Z}$ using
the variances of the posterior distributions.
Finally, we define a decision-maker by its disparity and the corresponding uncertainty, enabling an enhanced discrimination assessment.

\subsection{Estimating Treatment Probabilities}\label{section:beliefs}
Earlier, we defined treatment as the probability of group
$i \in g$ receiving some specific event $E_1$ given $E_2$.
Let us consider the hiring process as an example again, then
$P(E_1 \mid E_2, Z=i)$ could represent the chances of group $i$
receiving a job offer $E_1$ under the condition of having a certain qualification $E_2$.
This example depicts a Binomial distribution, where the outcome is binary.
When having samples from the hiring process, we can denote the number of
applicants in group $i$ as:
\begin{equation}\label{eq:ni}
	n_i = |\{Z = i\} \cap E_2|,
\end{equation}
and those of group $i$ who received a job offer as:
\begin{equation}\label{eq:ki}
	k_i = |E_1 \cap \{Z = i\} \cap E_2|.
\end{equation}

\subsubsection{Frequentist Approach}
In frequentist statistics, the probability of a
Binomial distribution is estimated using empirical counts\footnote{
Maximum likelihood estimation}.
For shorthand, let's denote $p_i \coloneqq P(E_1 \mid E_2, Z=i)$, then the estimate is given by:
\begin{equation}\label{eq:freqp}
	\hat{p_i} = \frac{|E_1 \cap \{Z = i\} \cap E_2|}{|\{Z = i\} \cap E_2|} = \frac{k_i}{n_i}.
\end{equation}
With more samples, the estimate becomes more accurate, i.e.,
$\lim_{{n_i \to \infty}} \hat{p_i} = p_i$.
In practice, $n_i$ can be small and therefore
the estimate $\hat{p_i}$ can be quite different from the
true probability $p_i$.

\subsubsection{Bayesian Approach}
In Bayesian statistics~\citep{gelman1995bayesian}, the quantification of uncertainty involves modeling
$p_i$ as a random variable rather than setting it to a
fixed constant as in Equation~\eqref{eq:freqp}.
We start with a \emph{prior distribution} $p(p_i)$ that represents
our beliefs before observing any data $\mathcal{D}$.
When estimating parameters for a Binomial event,
the Beta distribution,
denoted with $\mathcal{B}(\alpha, \beta)$, is commonly used as the
prior distribution~\citep{gelman1995bayesian}.
Similarly to the Binomial distribution, it models
binary outcomes.
It does this with two shape parameters, $\alpha$ and $\beta$.
To yield a non-informative uniform prior~\citep{gelman1995bayesian},
both parameters are usually set with
\begin{align}
	\alpha_{\text{prior}} &= 1, \\
	\beta_{\text{prior}} &= 1. \nonumber
\end{align}
This setting is motivated by the principle of indifference in Bayesian statistics
and aligns with Laplace's rule of succession.
In the next step, the prior distribution
\begin{equation}
	p(p_i) = \mathcal{B}(\alpha_{\text{prior}}, \beta_{\text{prior}})
\end{equation}
is updated.
The updated distribution is known as the \emph{posterior distribution}
$p(p_i | \mathcal{D})$, which models the distribution of $p_i$ after observing data $\mathcal{D}$
and represents our current beliefs.

According to \citet{gelman1995bayesian},
the posterior can be obtained by adding the corresponding
number of successes and failures to the shape parameters
of the prior distribution.
Specifically, the parameters for the posterior are:
\begin{align}
	\alpha^{(i)}_{\text{post.}} &= \alpha_{\text{prior}} + k_i, \\
	\beta^{(i)}_{\text{post.}} &= \beta_{\text{prior}} + n_i - k_i. \nonumber
\end{align}
With the posterior distributions:
\begin{equation}\label{eq:posterior}
	p(p_i | \mathcal{D}) = \mathcal{B}(\alpha^{(i)}_{\text{post.}}, \beta^{(i)}_{\text{post.}})
\end{equation}
for each group $i \in g$ in hand,
we can compare the group disparities more comprehensively.
The posterior distributions allow us to derive alternative
definitions for treatment and disparity.
Since $p_i$ and $p_j$ are not single probabilities under
this paradigm, the definitions of treatment
and disparity undergo notational modifications.

\begin{definition}[Bayesian Treatment]\label{eq:bayp}
	We denote the expected value of the posterior $p(p_i | \mathcal{D})$ as $\mathbb{E}(p_i | \mathcal{D})$.
	It is given by~\citep{gelman1995bayesian}:
	\begin{equation*}
		\mathbb{E}(p_i | \mathcal{D}) = \frac{\alpha^{(i)}_{\text{post.}}}{\alpha^{(i)}_{\text{post.}} + \beta^{(i)}_{\text{post.}}}
	\end{equation*}
	and is the Bayesian estimate of $p_i$.
\end{definition}

\begin{definition}[Bayesian Disparity]\label{def:bayesdisparity}
	Denoting $p_i$ and $p_j$ with the expected value,
	the Bayesian disparity $\delta_{Z}^{\text{(Bay.)}}$ becomes:
	\begin{equation*}
		\delta_{Z}^{\text{(Bay.)}}(i, j, E_1, E_2) = |\mathbb{E}(p_i | \mathcal{D}) - \mathbb{E}(p_j | \mathcal{D})|.
	\end{equation*}
\end{definition}

$\delta_{Z}^{\text{(Bay.)}}$ differs from $\delta_{Z}$ marginally
if the number of samples is small.
We leave the choice of the disparity definition to the user.
We suggest using Bayesian disparity, if there is an initial belief
that both groups have a 50\% chance of receiving the favorable outcome.
If such a belief is not present, the frequentist disparity is
more suitable and less biased.
We use the frequentist disparity in our work.

\subsection{Modeling (Un)certainty of (Un)fairness}
\begin{figure}[tb]
	\centering
	\includegraphics[width=0.48\textwidth]{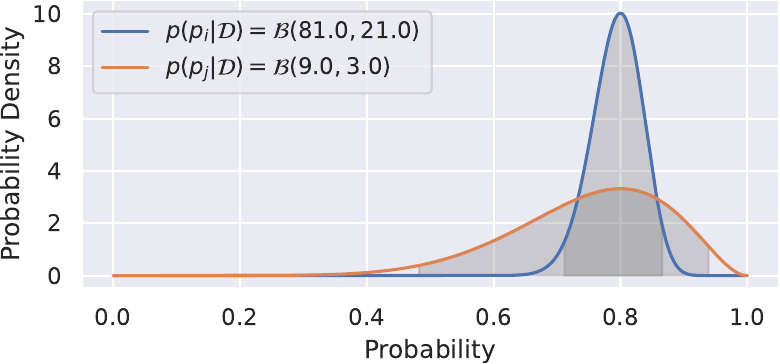}
	\caption{Group $i$ comprises $n_i=100$ individuals, with $k_i=80$ receiving the favorable outcome, while group $j$ consists of $n_j=10$ individuals, of which $k_j=8$ experience the favorable outcome. The figure displays the probability density functions of the posteriors. The filled areas mark the 95\% credible intervals of each distribution. Noticeable, we are less certain about the data from group $j$. In frequentist statistics, both groups are
		treated equally, but the Bayesian approach enables differentiating the groups.
	}
	\label{fig:posteriorexample}
\end{figure}

As seen in Figure~\ref{fig:posteriorexample}, even with same frequentist
treatments for group $i$ and $j$ (80\%), the posterior distributions
are vastly different.
This is due to the different group sizes $n_i$ and $n_j$
and is signified by the variances of the posteriors.
Hence, the variances of the posterior distributions
describe the underlying uncertainties.
We denote the variance with $\sigma^2_{\mathcal{B}}$ and it is defined by~\cite{gelman1995bayesian}:
\begin{equation}
	\sigma^2_{\mathcal{B}}(\alpha, \beta) = \frac{\alpha \beta}{(\alpha + \beta)^2 (\alpha + \beta + 1)}.
\end{equation}

Due to interpretability reasons, we aim to normalize the variance
to the closed interval $[0, 1]$, where 0 represents no uncertainty
and 1 represents maximum uncertainty.
For this, it is essential to consider a few characteristics
of the variance.
Notably, $\sigma^2_{\mathcal{B}}$ is monotonically decreasing
with respect to the shape parameters $\alpha$ and $\beta$,
i.e., larger parameters lead to a smaller variance. 
Given that these shape parameters are natural numbers,
the largest achievable variance of the posterior distribution,
derived from Equation~\eqref{eq:posterior}, is given by
$\sigma^2_{\mathcal{B}}(1, 2)$,
or equivalently $\sigma^2_{\mathcal{B}}(2, 1)$.
We employ this maximum variance as a scaling factor, resulting in the following normalized variance $\hat{\sigma}^2_{\mathcal{B}}$:
\begin{equation}\label{eq:bayesweightsnorm}
	\hat{\sigma}^2_{\mathcal{B}}(\alpha, \beta) \coloneqq \frac{\sigma^2_{\mathcal{B}}(\alpha, \beta)}{\sigma^2_{\mathcal{B}}(1, 2)}.
\end{equation}

When comparing the disparities between two groups $i, j \in g$,
we can use both normalized variances of the posteriors to obtain the uncertainty of the disparity and answer research question \textbf{RQ1}
with the following definition.

\begin{definition}[Uncertainty]\label{def:uncertainty}
	We define the uncertainty of the disparity between two groups $i, j \in g$ as the mean of the normalized variances of their posterior distributions:
	\begin{equation}
		\bar{\sigma}^2_{\delta_{Z}}(i, j, E_1, E_2) = \frac{\hat{\sigma}^2_{\mathcal{B}}(\alpha^{(i)}_{\text{post.}}, \beta^{(i)}_{\text{post.}}) + \hat{\sigma}^2_{\mathcal{B}}(\alpha^{(j)}_{\text{post.}}, \beta^{(j)}_{\text{post.}})}{2}. \nonumber
	\end{equation}
\end{definition}

By taking the average, the uncertainties from both groups are combined.
A higher uncertainty score indicates a lower precision of the disparity estimate and vice versa.
A maximum uncertainty of 1 is achieved if
both groups consist of a single individual.
We can now define a decision-maker by its disparity and the corresponding uncertainty in the following definition.

\begin{definition}[Decision-Maker]\label{def:decisionmaker}
	A decision-maker $D \in [0, 1]^2$ is defined by its disparity and the corresponding uncertainty:
	\begin{equation}
		D = (\delta_{Z}, \bar{\sigma}^2_{\delta_{Z}}). \nonumber
	\end{equation}
\end{definition}

\section{Ranking Decision-Makers}\label{section:method}
In this section, we begin by defining preferences over decision-makers,
establishing the criteria for what is deemed to be more or less fair.
Subsequently, we formulate a utility function that maps decision-makers to
values that represent the preferences and enables ranking, thus answering research question \textbf{RQ2}.
A higher utility value indicates a more preferred decision-maker.
To autonomously select the optimal decision-maker,
we iterate through all candidates to find the
decision-maker with the maximal utility value (\textbf{RQ3}).
Additionally, we introduce the concept of indifference curves,
offering insights into cases where two different decision-makers
are equally preferred.
The preference definitions in this section are mainly inspired by the
work of~\citet{levin2004socialchoice} and were adapted to fit our context.

\subsection{Preferences}
We recall the preferences~\eqref{eq:preferences1}-\eqref{eq:preferences3} from Section~\ref{section:introduction}
we have over decision-makers.
We first introduce the definition of a preference relation
and then define the preferences~\eqref{eq:preferences1}-\eqref{eq:preferences3} formally using the definition of a decision-maker.

\begin{definition}[Preference Relation]\label{def:preferencerelation}
	We denote a strict preference relation with
	$\succ$ or $\prec$ and write $D_1 \succ D_2$ to signify that
	decision-maker $D_1$ is preferred over $D_2$.
	The symbol $\sim$ denotes indifference, i.e., $D_1 \sim D_2$
	means that $D_1$ and $D_2$ are equally preferred.
	The strict preference relation is transitive,
	while the indifference relation is reflexive and transitive.
\end{definition}

\begin{definition}[Trivial Preferences]\label{def:preferences}
	We have following preferences over decision-makers:
	\begin{align}
		\text{fair certain} &\succ \text{fair uncertain}: &(0, 0) &\succ (0, 1) \nonumber \\
		\text{fair uncertain} &\succ \text{unfair uncertain}: &(0, 1) &\succ (1, 1) \nonumber \\
		\text{unfair uncertain} &\succ \text{unfair certain}: &(1, 1) &\succ (1, 0) \nonumber
	\end{align}

	Due to transitivity, we can derive additional preferences:
	\begin{align}
		\text{fair certain} &\succ \text{unfair uncertain}: &(0, 0) &\succ (1, 1) \nonumber \\
		\text{fair certain} &\succ \text{unfair certain}: &(0, 0) &\succ (1, 0) \nonumber \\
		\text{fair uncertain} &\succ \text{unfair certain}: &(0, 1) &\succ (1, 0) \nonumber
	\end{align}
\end{definition}

The listed preferences are trivial and extreme cases,
where a decision-maker is characterized by extreme instances of
(un)fairness and (un)certainty, i.e., $D \in \{0, 1\}^2$.
We note that listing all preferences over decision-makers,
as defined in Definition~\ref{def:preferences}, is impossible
because infinite decision-makers exist in the continuous space,
thus making the preference relation incomplete.
We call any preference that is not trivial a \emph{non-trivial preference}.

\begin{definition}[Non-Trivial Preference]\label{def:nontrivialpreferences}
	$D_1 \succ D_2$ is a non-trivial preference if and only if
	$D_1, D_2 \in ]0, 1[^2$.
\end{definition}

Modeling non-trivial preferences can be challenging as we are comparing
decision-makers that are neither extremely fair, unfair, certain, nor uncertain.
However, it is possible to infer non-trivial preferences
from the trivial ones, as we will show in the next section.

\subsection{Ranking with Utility Values}
By introducing a utility function $u$, we can translate preferences
over decision-makers into utility values that enable proper comparison, i.e.,
\begin{equation}
	D_1 \succ D_2 \implies u(D_1) > u(D_2).
\end{equation}
Importantly, the utility function must satisfy all trivial preferences
from Definition~\ref{def:preferences}.
However, this still leaves us open with infinitely many decision-makers
that are not covered by the defined preferences, specifically for any
$D \in ]0, 1[$.
Therefore, we need to define a utility function that is able to assign a value to all possible decision-makers.
By doing so, we can rank all decision-makers accordingly to the defined
preferences and the undefined, non-trivial preferences.
For the latter, we assume that these preferences can be implied
from the utility:
\begin{equation}
	D_1 \succ D_2 \impliedby u(D_1) > u(D_2).
\end{equation}

\begin{definition}[Utility Function]\label{def:utility}
	Let $\mathbb{D} = [0, 1]^2$ be the set of all decision-makers,
	a utility function $u: \mathbb{D} \rightarrow \mathbb{R}$ is total
	and must fulfill all preferences from Definition~\ref{def:preferences}, that is:
	\begin{align}
		u(0, 0) &> u(0, 1) \nonumber \\
		u(0, 1) &> u(1, 1) \nonumber \\
		u(1, 1) &> u(1, 0), \nonumber
	\end{align}
	including all derived preferences due to transitivity.
\end{definition}

By demanding totality, we ensure that the utility function
is able to assign a value to every decision-maker $D \in [0, 1]^2$.
A possible utility function is given by the following example.
\begin{example}[TOPSIS Utility]\label{eq:utility}
	Motivated by TOPSIS~\cite{hwang1981topsis},
	we calculate the utility of decision-makers based on their distances
	to the ideal solution $(0, 0)$ and the worst solution $(1, 0)$.
	Because utility is to be maximized, distances should
	be penalized accordingly. We define
	$u_\text{topsis}: [0, 1]^2 \rightarrow [-1, 1]$ with:
	\begin{align*}
		u_\text{topsis}(\delta_{Z}, \bar{\sigma}^2_{\delta_{Z}}) &= \norm{(\delta_Z, \bar{\sigma}^2_{\delta_{Z}}) - (1, 0)}_2 -  \norm{(\delta_Z, \bar{\sigma}^2_{\delta_{Z}}) - (0, 0)}_2 \\
		&= \sqrt{(\delta_Z - 1)^2 + (\bar{\sigma}^2_{\delta_{Z}})^2} - \sqrt{(\delta_Z)^2 + (\bar{\sigma}^2_{\delta_{Z}})^2}.
	\end{align*}
\end{example}

\begin{theorem}\label{thm:utility}
	$u_\text{topsis}$ is a utility function as it is total,
	fulfills all preferences from Definition~\ref{def:preferences},
	and preserves the transitive preferences.
\end{theorem}

\begin{proof}
	Trivial. $u_\text{topsis}$ is total by definition, i.e.,
	a value $u_\text{topsis}(D)$ exists for all $D \in \mathbb{D}$.
	Next, input the values from Definition~\ref{def:preferences}
	and show that all preferences including the transitive ones hold.
\end{proof}

The idea behind $u_\text{topsis}$ is that
the decision-maker that is closer to the ideal decision-maker $(0, 0)$
and farther away from the worst decision-maker $(1, 0)$
is rewarded with a higher utility value.
The utility function is not unique and can be replaced by any other function fulfilling
the requirements from Definition~\ref{def:utility}.
Since we modeled the utility function in Example~\ref{eq:utility}
to favor certainly fair decision-makers and disfavor certainly unfair ones,
we can be sure that any decision-maker with a higher utility value
is more preferred than any other by rational users that have the same
preferences as in Definition~\ref{def:preferences}.

Because a normalized score is more intuitive,
stakeholders might prefer to use the utility function
from the following example.

\begin{example}[Normalized Utility]\label{eq:normalizedutility}
	We define a normalized utility function
	$u_\text{norm}: [0, 1]^2 \rightarrow [0, 1]$ with:
	\begin{equation}
		u_\text{norm}(\delta_{Z}, \bar{\sigma}^2_{\delta_{Z}}) = \frac{u_\text{topsis}(\delta_{Z}, \bar{\sigma}^2_{\delta_{Z}}) + 1}{2}.
	\end{equation}
\end{example}

\begin{theorem}
	$u_\text{norm}$ is a utility function as it is total,
	fulfills all preferences from Definition~\ref{def:preferences},
	and preserves the transitive preferences.
\end{theorem}

\begin{proof}
	Trivial. Apply the same steps as in the proof of Theorem~\ref{thm:utility}.
\end{proof}

\subsection{Objective Function and Selecting Optimal Decision-Maker}
Let us have a set of decision-makers $D = \{D_1, D_2, \ldots, D_m\}$,
then the approach to choose the optimal decision-maker $D^*$ is given by
solving the following optimization problem:
\begin{equation}
	D^* = \argmax_{D_i \in D} \quad u(D_i). \nonumber
\end{equation}
For a finite set of decision-makers, this can be solved efficiently
with brute-force search in $\mathcal{O}(m)$.

\subsection{Indifference Curve}
When two decision-makers have the same utility,
they are indifferent to each other, i.e., $D_1 \sim D_2$.
In such cases, the user is left with free choices to select
their optimal decision-maker.
All points having the same utility value lie on an indifference curve.
It can be derived by solving the following equation:
\begin{equation}\label{eq:indifferencecurve}
	u(D_1) = u(D_2).
\end{equation}
Let us denote $D_1 = (a_1, a_2)$, $D_2 = (b_1, b_2)$,
then we specifically solve:
{\small
\begin{equation}
	\sqrt{(a_1 - 1)^2 + a_2^2} - \sqrt{a_1^2 + a_2^2} = \sqrt{(b_1 - 1)^2 + b_2^2} - \sqrt{b_1^2 + b_2^2}. \\
\end{equation}
}

Depending on which variable ($a_1, a_2, b_1, b_2$) is treated as a constant, the analytical
solution can become excessively long.
We did find such solutions for the indifference curve with
symbolic computation~\citep{sympy2017}, but they are not insightful.
We found a trivial solution with:
\begin{equation}
	u(D_1) = u(D_2) = 0.
\end{equation}
For this case, the curve is given when $a_1 = b_1 = 0.5$
and $a_2, b_2$ can be any value in $[0, 1]$.
This means that decision-makers are indifferent as long as 
their disparities are both 50\%.
Utility values are also negative if the disparity is higher than 50\%
and positive if it is lower.

\section{Experiments}
Before diving into the experiments, we revisit the example from
Figure~\ref{fig:motivation}.
We calculate the disparity and uncertainty for the two recruiters,
$A$ and $B$, and list the utility values using $u_\text{topsis}$ in Table~\ref{table:resultmotivation}.
When comparing the disparities, both recruiters are indifferent as they
are equally unfair towards group $j$.
According to the utility values, recruiter $B$ has a higher utility than $A$
and is therefore more preferred.
This aligns with the intuition that we are more uncertain about
$B$'s unfairness than $A$'s.

\begin{table}[htb]
	\centering
	\caption{Revisiting example given in Figure~\ref{fig:motivation}.}
	\label{table:resultmotivation}
	\resizebox{\columnwidth}{!}{%
	\begin{tabular}{lllllllll}
	\toprule
	Recruiter & $n_i$ & $k_i$ & $n_j$ & $k_j$ & $\hat{p}_i$ & $\hat{p}_j$ & DM $(\delta_{Z}, \bar{\sigma}^2_{\delta_{Z}})$ & Utility \\
	\midrule
	$A$ & 3 & 3 & 3 & 0 & 100\% & 0\% & (1.000, 0.480) & -0.629 \\
	$B$ & 1 & 1 & 1 & 0 & 100\% & 0\% & (1.000, 1.000) & -0.414 \\
	\bottomrule
	\end{tabular}
	}
\end{table}

To explore our methodology more extensively,
we conduct experiments on synthetic and real-world datasets.
We use synthetic data to have full control over
the disparities and uncertainties of decision-makers.
This is done by setting different group treatments
and varying the group sizes.

\subsection{Synthetic Data}
\begin{table}[tb]
	\centering
	\caption{Four decision-makers with the highest and lowest utility values from the synthetic data created in the experiments.}
	\label{table:decisionmakers}
	\resizebox{\columnwidth}{!}{%
	\begin{tabular}{lllllllll}
	\toprule
	Rank & $n_i$ & $k_i$ & $n_j$ & $k_j$ & $\hat{p}_i$ & $\hat{p}_j$ & DM $(\delta_{Z}, \bar{\sigma}^2_{\delta_{Z}})$ & Utility \\
	\midrule
	1 & 50 & 50 & 50 & 50 & 100\% & 100\% & (0.000, 0.006) & 0.994 \\
	2 & 50 & 0 & 50 & 0 & 0\% & 0\% & (0.000, 0.006) & 0.994 \\
	3 & 50 & 49 & 50 & 49 & 98\% & 98\% & (0.000, 0.013) & 0.988 \\
	4 & 50 & 1 & 50 & 1 & 2\% & 2\% & (0.000, 0.013) & 0.988 \\
	\cline{1-9}
	4897 & 50 & 0 & 50 & 49 & 0\% & 98\% & (0.980, 0.009) & -0.958 \\
	4898 & 50 & 49 & 50 & 0 & 98\% & 0\% & (0.980, 0.009) & -0.958 \\
	4899 & 50 & 50 & 50 & 0 & 100\% & 0\% & (1.000, 0.006) & -0.994 \\
	4900 & 50 & 0 & 50 & 50 & 0\% & 100\% & (1.000, 0.006) & -0.994 \\
	\bottomrule
	\end{tabular}
	}
\end{table}

We first generate group sizes $(n_i, n_j) \in \{1, 5, 10, 50\}^2$.
Each group $i \in g$ can receive any number of favorable outcomes $k_i$
based on its size $n_i$.
For example, if $n_i = 5$, then $k_i$ can be any natural number in $[0, 5]$.
Decision-makers are then created by calculating the disparity and uncertainty
through all possible combinations of group sizes and treatments.
This results in 4\,900 decision-makers.
We then calculate the utility value using $u_\text{topsis}$ for each decision-maker.

We list four decision-makers with the highest and lowest utility values
from the synthetic data in Table~\ref{table:decisionmakers}.
The most favorable decision-makers, with the same highest utility values,
are those where all individuals from both groups
either receive the favorable or unfavorable outcome, i.e.,
$k_i, k_j \in \{0, n_i\}$ with $k_i = k_j$.
Groups are essentially treated equally and consist of large sample sizes.
The least favorable decision-makers are the ones, where the disparity is
maximized and the uncertainty is lowest.
This aligns with the intuition that decision-makers,
where we know that they are without a doubt unfair, are less preferred.

\subsection{COMPAS Dataset}
We use the COMPAS~\cite{larsonangwinmattukirchner2016} dataset
to evaluate decision-makers.
The dataset contains information
about defendants and their criminal histories.
We compare different machine learning models, namely
\emph{Logistic Regression} (LR),
\emph{Support Vector Machine} (SVM),
\emph{Random Forest} (RF), and \emph{k-Nearest Neighbors} (KNN),
that predict whether a defendant will be rearrested within two years.
These models act as decision-makers in our context.
The dataset consists of 7\,214 samples, and
we use an 80/20 split for training and testing.
Different from the processed versions of COMPAS in
other fairness libraries~\cite{aif360,fairlearn},
the protected attribute `\emph{race}'
has not been reduced to two categories but is utilized
in its original form.
To calculate the disparity for this,
we report the following difference~\cite{liobait2017MeasuringDI,fairlearn}:
\begin{equation}
	\delta_{Z} = \max_{i \in g} P(\hat{Y}=0 \mid Z=i) - \min_{j \in g} P(\hat{Y}=0 \mid Z=j),
\end{equation}
where $\hat{Y} = 0$ is the predicted outcome on the test set,
noting that it is considered the
favorable outcome as it indicates that a defendant
will not be rearrested.
Using this formula, the most and least privileged groups
can differ for each model.

\begin{table*}[htb]
	\centering
	\caption{Results from the COMPAS dataset.}
	\label{table:resultscompas}
	\resizebox{\textwidth}{!}{%
	\begin{tabular}{llllllllllll}
	\toprule
	Model & Most Privileged ($i$) & Least Privileged ($j$) & $n_i$ & $k_i$ & $n_j$ & $k_j$ & $\hat{p}_i$ & $\hat{p}_j$ & DM $(\delta_{Z}, \bar{\sigma}^2_{\delta_{Z}})$ & Utility & Accuracy \\
	\midrule
	LR & Asian & Native American & 6 & 6 & 4 & 2 & 100\% & 50\% & (0.500, 0.431) & 0	& 72\% \\
	KNN & Asian & Native American & 6 & 5 & 4 & 0 & 83.33\% & 0\% & (0.833, 0.366) & -0.508 & 66.81\% \\
	SVM & Asian & Native American & 6 & 6 & 4 & 0 & 100\% & 0\% & (1.000, 0.288) & -0.753 & 71.10\% \\
	RF & Asian & Native American & 6 & 6 & 4 & 0 & 100\% & 0\% & (1.000, 0.288) & -0.753 & 70.20\% \\
	\bottomrule
	\end{tabular}
	}
\end{table*}

Table~\ref{table:resultscompas} displays the results of the
experimentation on the COMPAS dataset.
The models are ranked based on their utility values with $u_\text{topsis}$.
We also report the accuracy of each model.
The Logistic Regression model has the highest utility value and is therefore the most preferred.
Interestingly, we observed that Asians are always the most privileged group, while Native Americans are always the least privileged group.
Nearly all Asians receive a favorable outcome,
while only a few Native Americans do.
Both groups come with a small sample size and are therefore associated with high uncertainty.
In this real-world scenario, ranking models by their utility values aligns
with ranking them by the disparity $\delta_Z$.
This is because the utility function is designed to favor decision-makers
with lower disparities.
However, utility values contain information about the uncertainty of the disparities.
Moreover, as illustrated in the example from Table~\ref{table:resultscompas}, utility values are essential for distinguishing between decision-makers who exhibit the same level of disparity.
In cases where both disparity and uncertainty are equal,
the utility values are also the same.
This is the case for the SVM and RF models in our experiment.
For this, we advise considering the accuracy of the models as well.
To conclude, LR has the highest utility value and accuracy,
making it the most suitable model for recidivism prediction in this case.

\subsection{Summary of Results}
Our work addresses three key research questions. Firstly, we establish a method to distinguish between decision-makers exhibiting the same levels of discrimination
by integrating uncertainty into our analysis (\textbf{RQ1}).
This involves modeling the uncertainty of the measured disparity of outcomes
between groups.
Using both disparity and uncertainty, we define a decision-maker
and establish our preferences among them.
Secondly, to compare decision-makers within the continuous space of preferences, we introduce a utility function that evaluates each candidate.
The utility values are then used to rank all decision-makers according to the defined preferences (\textbf{RQ2}).
Lastly, to identify the optimal decision-maker, we introduce an optimization objective, allowing us to select the most suitable candidate,
thus addressing \textbf{RQ3}.
The synthetic and real-world experiments demonstrate the practical usability
and necessity of our methodology to reliably assess the fairness of decision-makers.
\section{Discussion}\label{section:discussion}
While we answered all research questions prior, we want to discuss several aspects 
of our methodology, including the scope of our work, in this section.

It is important to model the utility function in such a way that it reflects
the user's preferences.
This is because non-trivial preferences are implied
by the utility function.
Here, we refer the reader to methods that map multiple criteria to a single value,
such as TOPSIS~\cite{hwang1981topsis} or the
Analytic Hierarchy Process~\cite{saaty1984ahp}.
Ranking decision-makers based on the utility function is a good starting point
to check if the preferences are correctly modeled.

Another important aspect is the indifference curve.
We found that decision-makers are indifferent to each other
as long as their disparities are both 50\%.
Here, the utility function is not sufficient to differentiate between
decision-makers, and the choice is left to the user.
We discourage choosing such a decision-maker where the
uncertainty is close to zero. This is because 50\% disparity is quite high in practice.
Decision-makers with a higher level of uncertainty are more preferred in such cases.

Our methodology is not invulnerable to manipulation.
For example, if a human decision-maker is aware of
the internal workings of our method,
he or she could
artificially increase the uncertainty of their disparity to appear
less discriminatory. In a hiring scenario, this can be done by
generally rejecting candidates coming from a very marginalized
group where the number of samples is small.
In such a case, minority groups should be grouped together into
one large group to avoid this kind of manipulation.

\section{Conclusion}
When dealing with small sample sizes, particularly in the case of
minority groups,
we are often uncertain about the collected data and the
information derived from it.
Group fairness metrics aim to report how different
groups are treated based on some specified events and outcomes,
disregarding uncertainty.
Therefore, we first introduce a method utilizing Bayesian statistics
to quantify the uncertainty of the disparity of group treatments and
employ them to enhance the assessment of discrimination.
With both the disparity and the uncertainty, we define decision-makers and derive preference relations over them.
By introducing a utility function that aligns with these preferences and is
defined for every possible decision-maker, we are able to select the most
preferred decision-maker with the largest utility from a set of candidates
using brute-force.
Our methodology comes with proven guarantees, and we have demonstrated its behavior on synthetic and real-world datasets.

The implications of our work are noteworthy, as we are able to differentiate
between systematic discrimination and random outcomes
and have defined preferences in such cases.
Decision-makers exhibiting discrimination on fewer samples are more preferred than those exhibiting discrimination on larger sample sizes.
Similarly, a certainly fair decision-maker is preferred over an uncertainly fair decision-maker. The latter is when the decision-maker receives fewer samples.
Our methodology can be used for a wide range of
applications, including evaluating machine learning models as well as
hiring and admission processes at companies and universities.
Additionally, the utility function can also be incorporated into the loss function
of a machine learning model to penalize decisions that are certainly unfair.

\section*{Ethics Statement}
With our proposed utility score, we address the issue of reporting discrimination
in uncertain cases. The proposed score can protect decision-makers from
discrimination accusations when the disparity they exhibited is uncertain,
while also ensuring that those who are clearly discriminatory are appropriately penalized.
Consequently, the societal impact of our work is positive.
Still, further research is needed to investigate the impact of our
method on several real-world applications.






\bibliography{references}

\end{document}